\RequirePackage{amsmath}         % Amsmath package for maths and equations

\documentclass[runningheads,a4paper]{llncs}

\usepackage{graphicx}            % For pdf/bitmapped graphics files
\usepackage{tikz}
\usepackage{amssymb}             % Maths symbols
\usepackage{amsfonts}            % To get the \mathbb alphabet
\usepackage{enumitem}            % For customisation of enumerate environments
\usepackage{multirow}            % For multi-row cells in tables
\usepackage{siunitx}             % For SI units
\usepackage{url}                 % For URL typesetting and line breaking
\usepackage{xspace}              % For smart spaces in commands
\usepackage[T1]{fontenc}         % For special characters such as '_'
\usepackage{hyperref}            % For hyperlinks
\usepackage{wrapfig}
\usepackage{todonotes}
\usepackage{booktabs}
\usepackage{caption}
\usepackage{subcaption}
\usepackage{xcolor}
\usepackage{array}
\newcolumntype{P}[1]{>{\centering\arraybackslash}p{#1}}

\graphicspath{{images/}}
\DeclareGraphicsExtensions{.pdf,.png,.jpg,.jpeg}

\newcommand{\Loss}{\mathcal{L}}

\newcommand{\seclabel}[1]{\label{sec:#1}}

\newcommand{\eqnlabel}[1]{\label{eqn:#1}}

\newcommand{\figref}[1]{Fig.~\ref{fig:#1}\xspace}

\newcommand{\cmnew}{CM740\xspace}

\usepackage[firstpage=true]{background}
\newcommand\copyrighttext{%
	\parbox{\textwidth}{
		\centering
		\footnotesize
		Accepted: RoboCup 2023: Robot World Cup XXVI. LNCS, Springer, to appear 2024.
	}
}

\SetBgContents{\copyrighttext}
\SetBgScale{1}
\SetBgColor{black}
\SetBgAngle{0}
\SetBgOpacity{1}
\SetBgPosition{current page.north}
\SetBgVshift{-2.5cm}
\SetBgHshift{3mm}
\begin{document}

% Start of the main matter
\mainmatter

% Paper title
\title{RoboCup 2023 Humanoid AdultSize Winner NimbRo: NimbRoNet3 Visual Perception and Responsive Gait with Waveform In-walk Kicks}
\titlerunning{RoboCup 2023 Humanoid AdultSize Winner NimbRo}

% Authors
\author{Dmytro Pavlichenko, Grzegorz Ficht, Angel Villar-Corrales, Luis Denninger, Julia Brocker, Tim Sinen, Michael Schreiber, and Sven Behnke}
\authorrunning{D. Pavlichenko, G. Ficht, A. Villar-Corrales, L. Denninger, J. Brocker, et al.}

% Institutes
\institute{Autonomous Intelligent Systems, Computer Science Institute VI, University of Bonn,\\ Germany,
\url{https://ais.uni-bonn.de},
\url{pavlichenko@ais.uni-bonn.de},
}

% Typeset the paper title
\maketitle

% Abstract
\begin{abstract}
The RoboCup Humanoid League holds annual soccer robot world championships towards the long-term objective of winning against the FIFA world champions by 2050.
The participating teams continuously improve their systems. 
This paper presents the upgrades to our humanoid soccer system, leading our team NimbRo to win the Soccer Tournament in the Humanoid AdultSize League at RoboCup 2023 in Bordeaux, France.
The mentioned upgrades consist of: an updated model architecture for visual perception, extended fused angles feedback mechanisms and an additional COM-ZMP controller for walking robustness, and parametric in-walk kicks through waveforms.
\end{abstract}

% Sections
\section{Introduction}

\begin{figure}[!b]
	\centering
	\includegraphics[height=61mm]{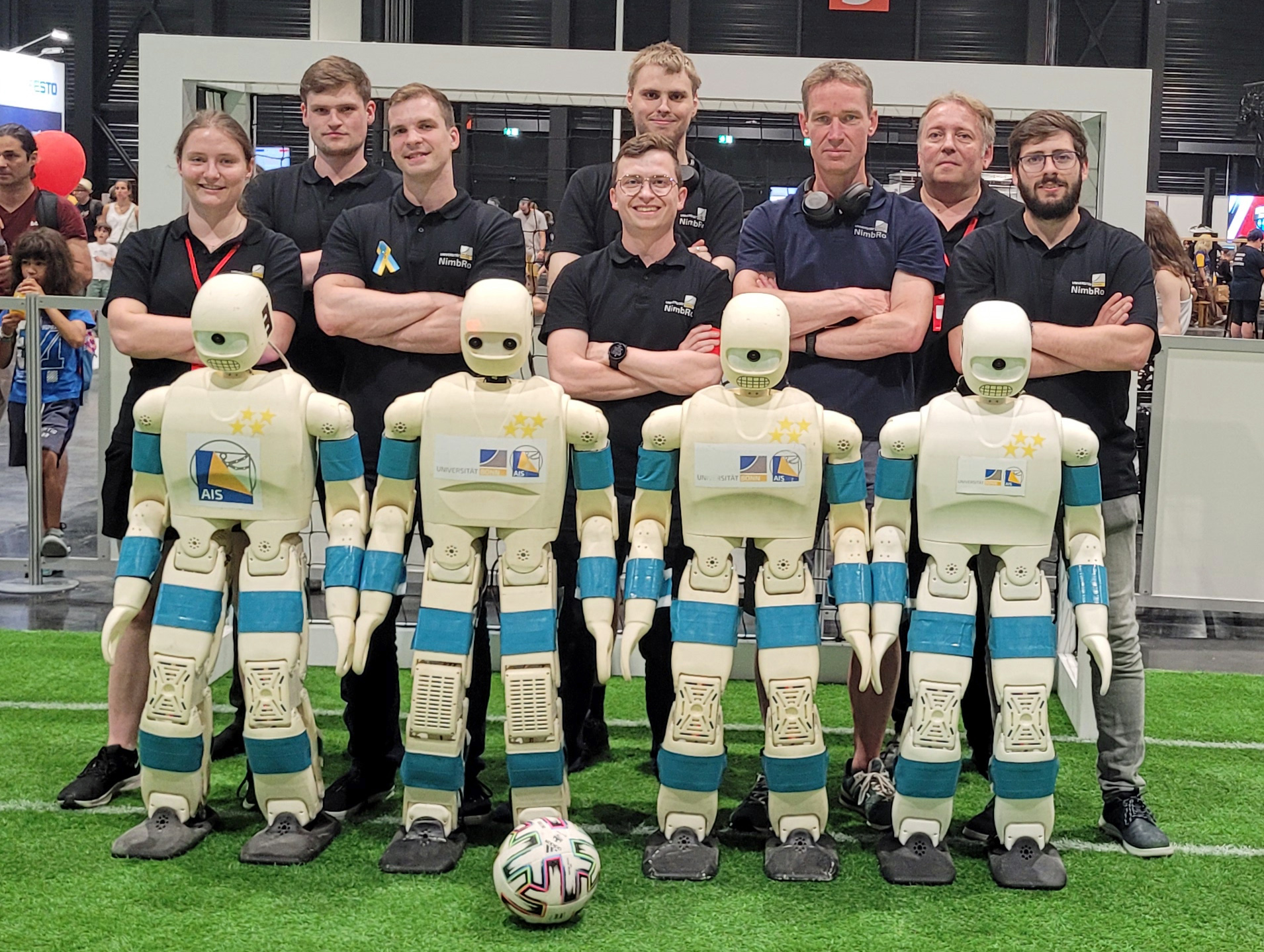} 
	\includegraphics[height=61mm]{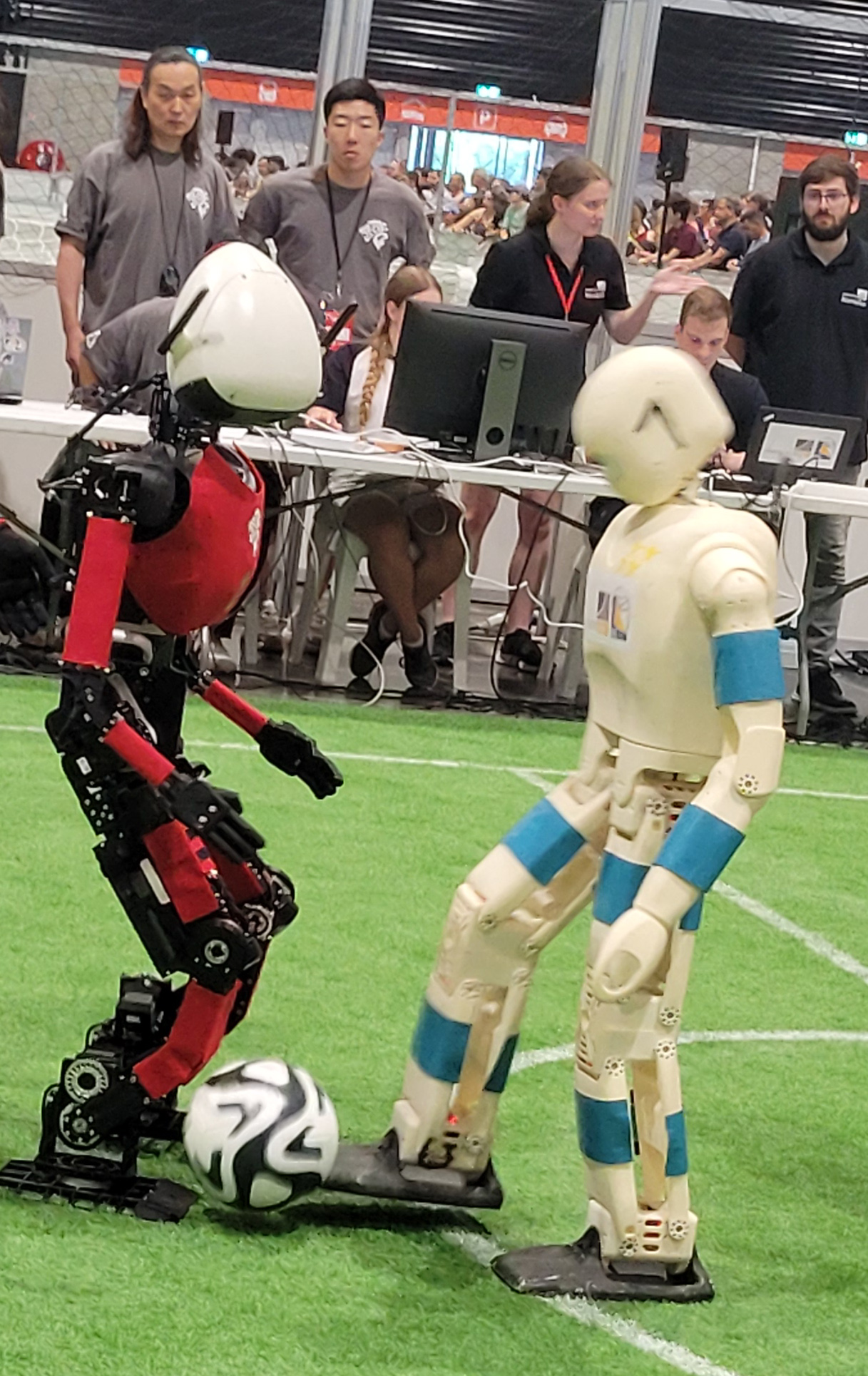}	
	\vspace*{-1ex}
	\caption{
		RoboCup 2023 in Bordeaux. Left: Team NimbRo AdultSize with NimbRo-OP2(X) robots, Right: Scene from the Final soccer game vs. HERoEHS (Korea). 
	}
	\label{fig:team}
\end{figure}

The Humanoid AdultSize League actively pursues the vision of RoboCup, i.e. by 2050 fully autonomous humanoid robot soccer players shall win against the FIFA world champion.
Participating teams continuously improve their systems towards this ambitious objective. 

For RoboCup 2023, we upgraded our perception pipeline with a new model capable of opponent robot pose estimation. Further, we extended the fused angle feedback mechanism for footstep adjustment and introduced an additional COM-ZMP controller to facilitate walking with higher stability. A novel gait based on a five-mass control principle was introduced for the push recovery technical challenge, improving resilience against pushes in the sagittal plane. Finally, we developed a parametric in-walk kick formulation through waveforms which enabled stronger kicks and easier tuning. 

These updates lead to an improved performance at RoboCup 2023, compared to the previous year~\cite{nimbro_winners_2022} and resulted in our team, shown in \figref{team}, winning the AdultSize soccer competition of the Humanoid League. A video of the highlights of our performance during RoboCup 2023 is available online\footnote{RoboCup 2023 NimbRo highlights video: \url{https://youtu.be/hKLC0Vz1GmM}}.

\section{3D-printed NimbRo-OP2(X) Humanoid Soccer Robots}
\seclabel{robot_platforms}

For the 2023 competition held in Bordeaux, we continued using our well-esta\-blished NimbRo-OP2~\cite{ficht2017nop2} and NimbRo-OP2X~\cite{ficht2018nimbro,ficht2020nimbro} humanoid robot platforms. 

The robots possess 18 Degrees of Freedom (DoF), with 5\,DoF per leg, 3\,DoF per arm, and 2\,DoF in the neck, driven by a total of 34 Robotis Dynamixel actuators. 
Their 4-bar leg parallel linkage increases stiffness in the 
sagittal plane---at the cost of locking the pitch of the foot relative to the trunk~\cite{ficht2021bipedal}. Synchronized 
actuation of the parallelogram with multiple actuators increases joint torques. 

Our robots are equipped with a strong Intel quadcore CPU with Nvidia GPU and a \cmnew microcontroller board,  
with a 6-axis IMU (3-axis accelerometer \& 3-axis gyro).
Their control architecture is based on ROS. By using off-the-shelf components, a single type of actuator, and 3D printing, the robots are affordable and their maintenance is significantly easier
than that of custom solutions~\cite{ficht2021bipedal}. 

Both the NimbRo-OP2 and NimbRo-OP2X humanoid soccer robot are open-source in terms of hardware\footnote{NimbRo-OP2X hardware: \url{https://github.com/NimbRo/nimbro-op2}} 
and software\footnote{NimbRo-OP2X software: \url{https://github.com/AIS-Bonn/humanoid_op_ros}}, with detailed documentation.

\section{Deep-learning based Visual Perception: NimbRoNet3}
\seclabel{perception}

\begin{figure}[t!]
	\centering
	\includegraphics[width=1.0\linewidth]{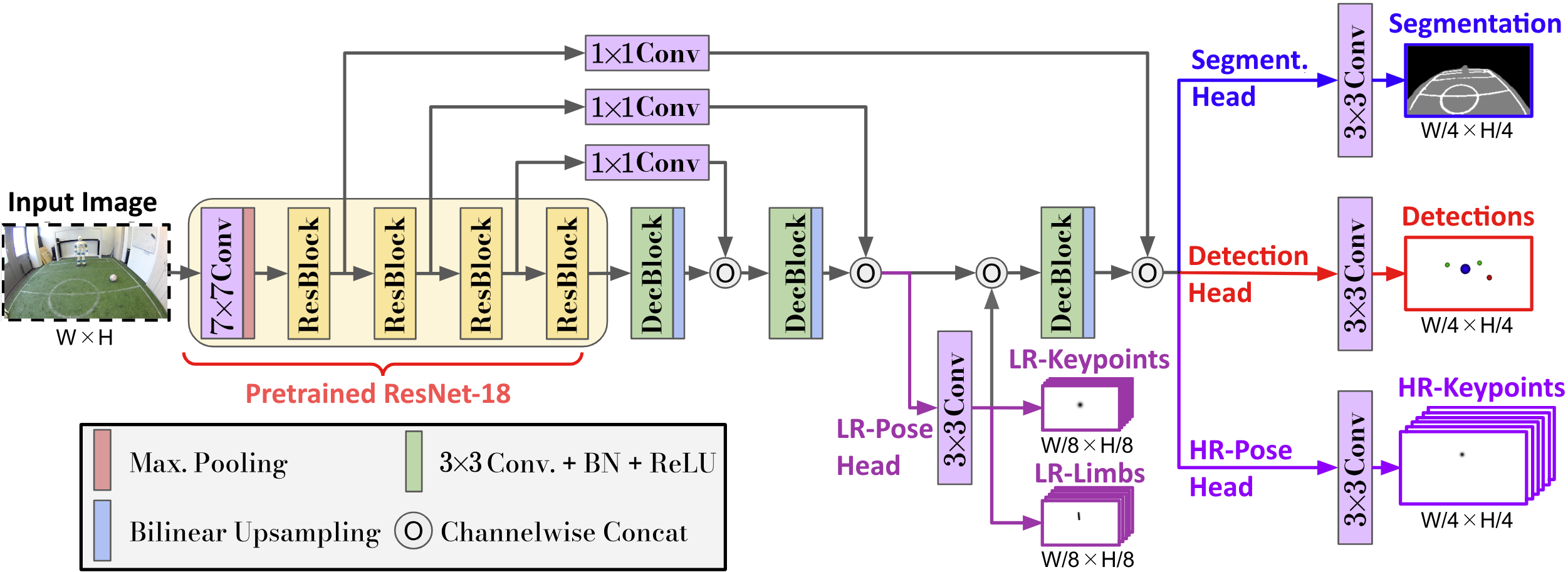}
	\caption{
		NimbRoNet3 model.
		Our model employs an encoder-decoder architecture with a pretrained ResNet-18 backbone, and four different network heads for \textcolor{blue}{\textbf{field segmentation}},
		\textcolor{red}{\textbf{object detection}}, and
		\textcolor{violet}{\textbf{HR/LR robot pose estimation}}.
	}
	\label{fig:NimbRoNet3}
	\vspace*{-4ex}
\end{figure}

Our humanoid NimbRo robots perceive the environment using a single 5\,MP Logitech C930e camera equipped with a wide-angle lens, providing a wide field-of-view in high-resolution.

To successfully perceive the environment in progressively larger soccer fields and more challenging game situations, we enhance our visual perception pipeline with respect to our previous NimbRoNet2~\cite{nimbro_winners_2019} model.
Our improved NimbRoNet3 perception model, depicted in Figure~\ref{fig:NimbRoNet3}, is a new convolutional neural network that simultaneously detects objects relevant to the soccer game, including the ball, goalposts, and robots; segments the field boundaries and lines; and estimates the pose of nearby robots.
Our visual perception pipeline, including a forward-pass through the model and post-processing steps, processes high-resolution input images ($3 \times 540 \times 960$) at a rate of 33.3 fps on the robot hardware, thus providing for real-time perception of the game environment.

Inspired by our previous works~\cite{nimbro_winners_2019,nimbro_winners_2022}, our perception model is a deep convolutional neural network employing an asymmetric encoder-decoder architecture with three skip connections, similar to UNet~\cite{unet2015}.
To minimize annotation efforts, we utilize a pretrained ResNet-18~\cite{resnet2016} backbone as encoder, after removing the final fully-connected and global average pooling layers.
Our decoder is formed by three convolutional blocks followed by bilinear upsampling, in contrast to the transposed convolution blocks employed in our previous work~\cite{nimbro_winners_2019}. Additionally, our decoder produces outputs with 1/4th of the input resolution, thus drastically reducing the number of required operations.
Three different skip connections bridge from the encoder to the decoder, thus maintaining high-resolution information, improving the spatial precision of the outputs~\cite{unet2015}. These skip connections contain a 1$\times$1 convolution to align the number of feature maps.

Our model is simultaneously trained to perform  object detection, field segmentation, and robot pose estimation using four distinct output heads:

\noindent$\bullet$\,\textbf{Object Detection:} The detection head predicts the location of goalposts, soccer balls and robots, represented by Gaussian heatmaps centered at the ball center and bottom-middle point of goalposts and robots. The exact object locations are retrieved from the predicted heatmaps in a post-processing step.
/

\noindent$\bullet$\,\textbf{Field Segmentation:} The segmentation head predicts a semantic category, e.g. background, field or line, for every pixel of the input image in order to obtain the segmentation of the soccer field.

\noindent$\bullet$\,\textbf{Robot Pose Estimation:} Inspired by recent work on human- and robot-pose estimation~\cite{higherhrnet,amini2022humanoid}, our model employs two convolutional output heads operating at different spatial resolution for the task of robot pose estimation.
To estimate the robot poses, our model predicts six different keypoints (i.e. head, trunk, left hand, right hand, left foot, and right foot) at high and low resolutions. Additionally, following~\cite{amini2022humanoid}, the low-resolution pose head also predicts the limbs connecting such keypoints.
During post-processing, the predicted keypoints and limbs are associated into robot poses following a greedy matching algorithm~\cite{openpose}.

\begin{figure}[t!]
	\centering \footnotesize \sf
	% row 1
	\includegraphics[width=.245\textwidth]{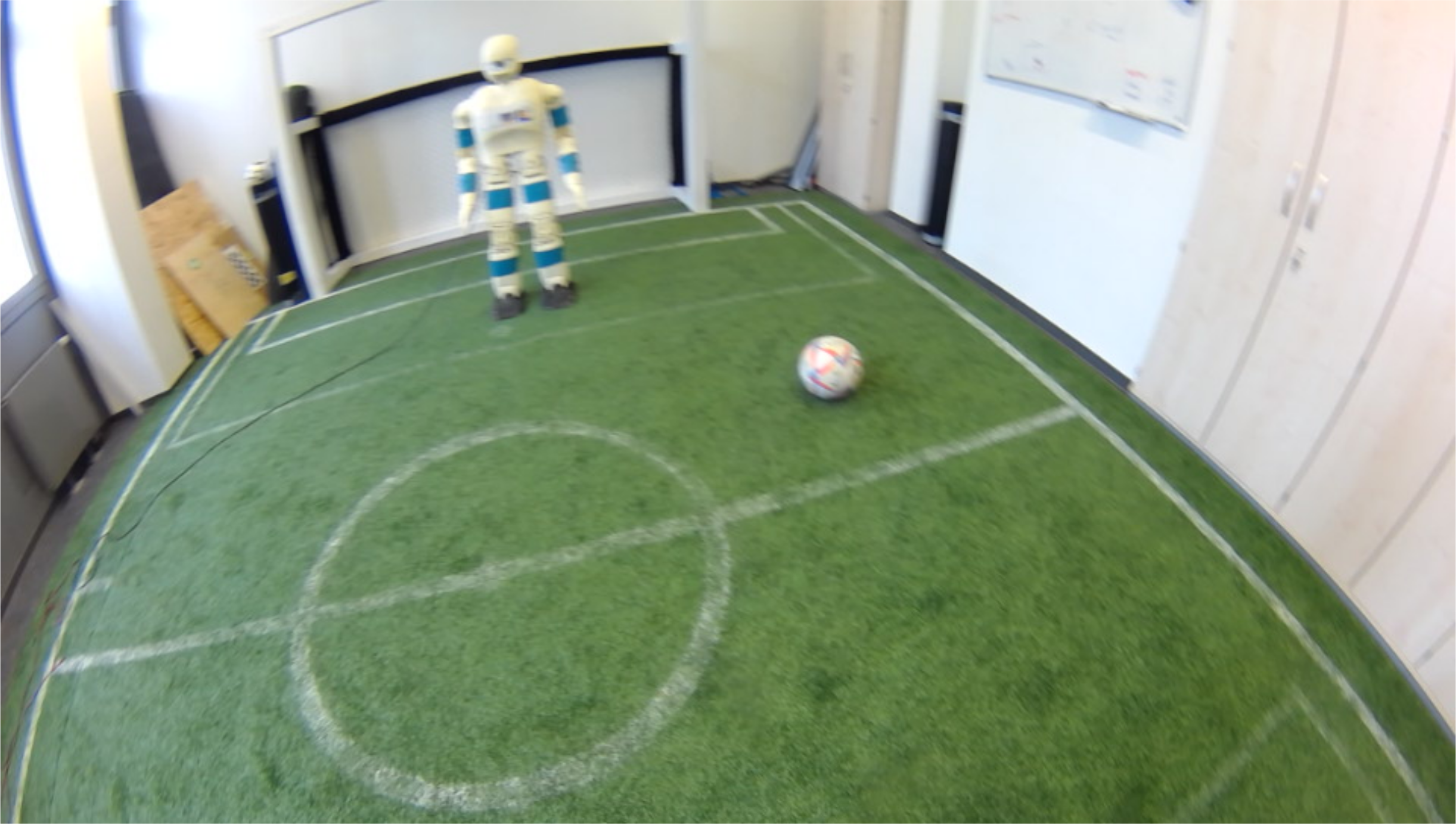}\,%	
	\includegraphics[width=.245\textwidth]{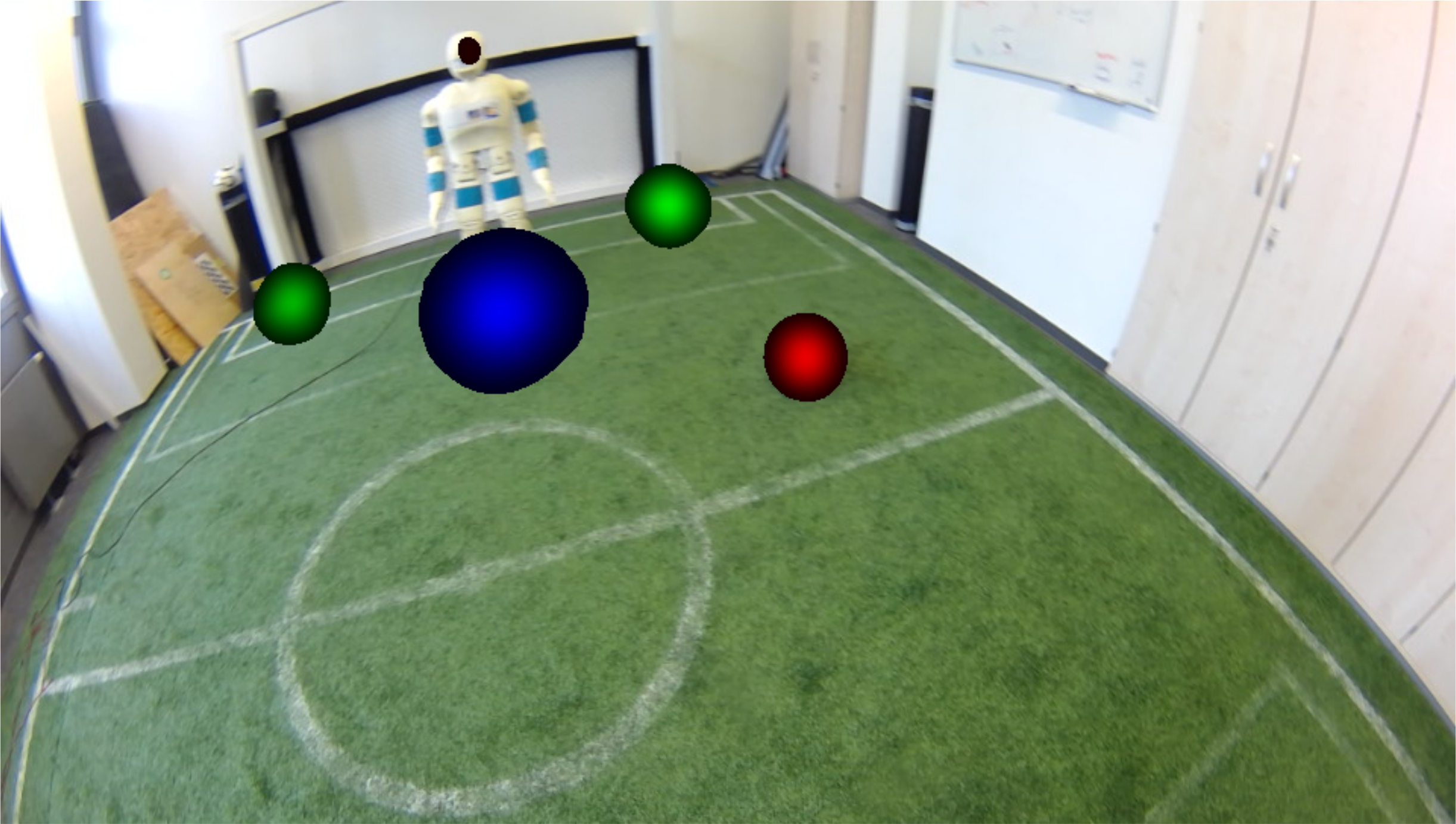}\,%
	\includegraphics[width=.245\textwidth]{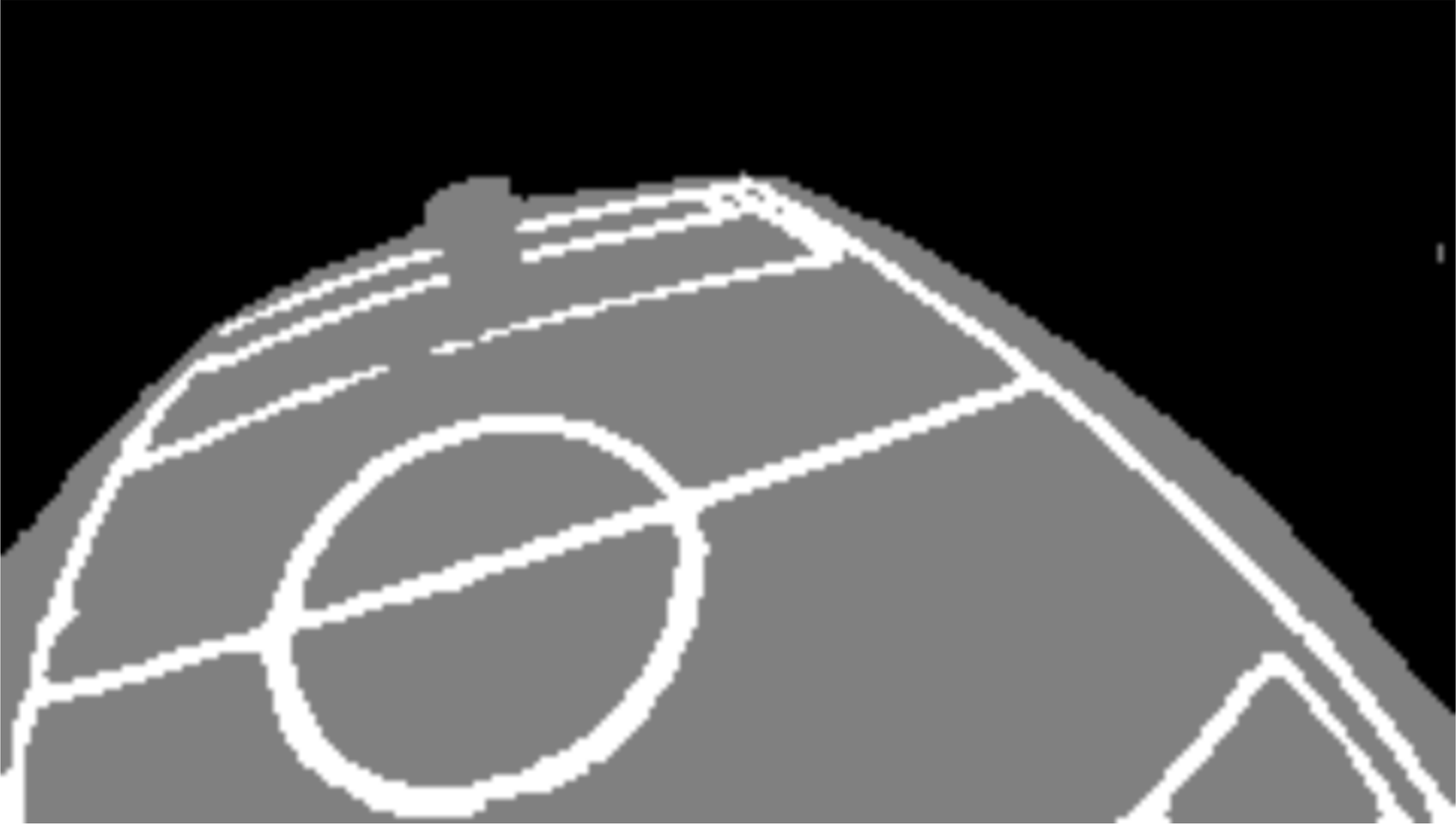}\,%
	\includegraphics[width=.245\textwidth]{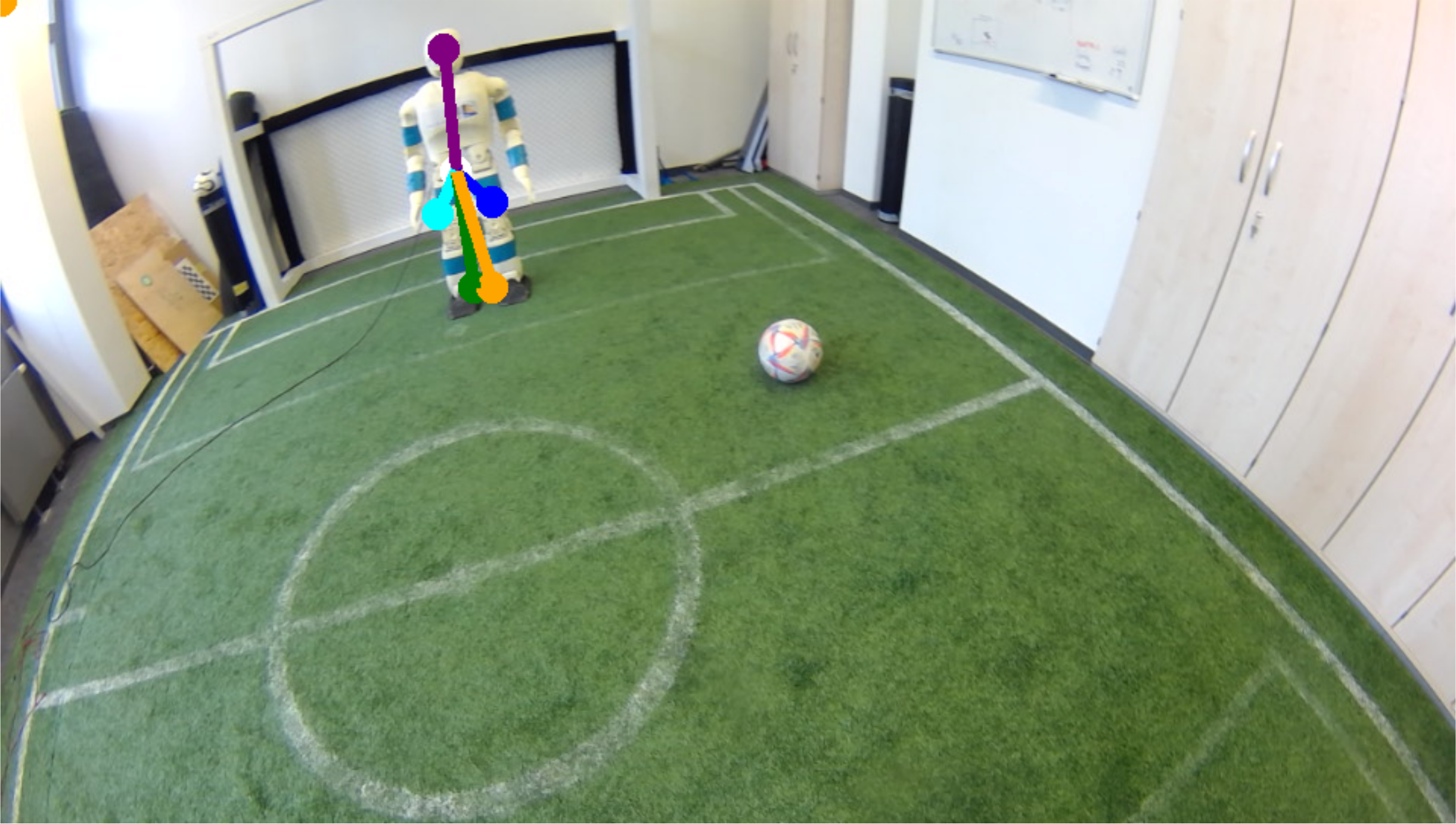}\vspace*{.1ex}\\
	\includegraphics[width=.245\textwidth]{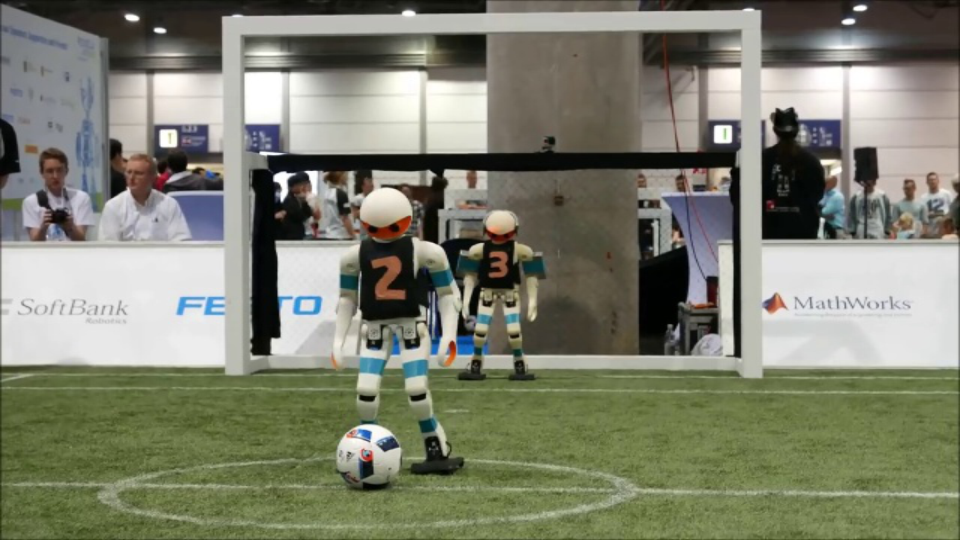}\,%
	\includegraphics[width=.245\textwidth]{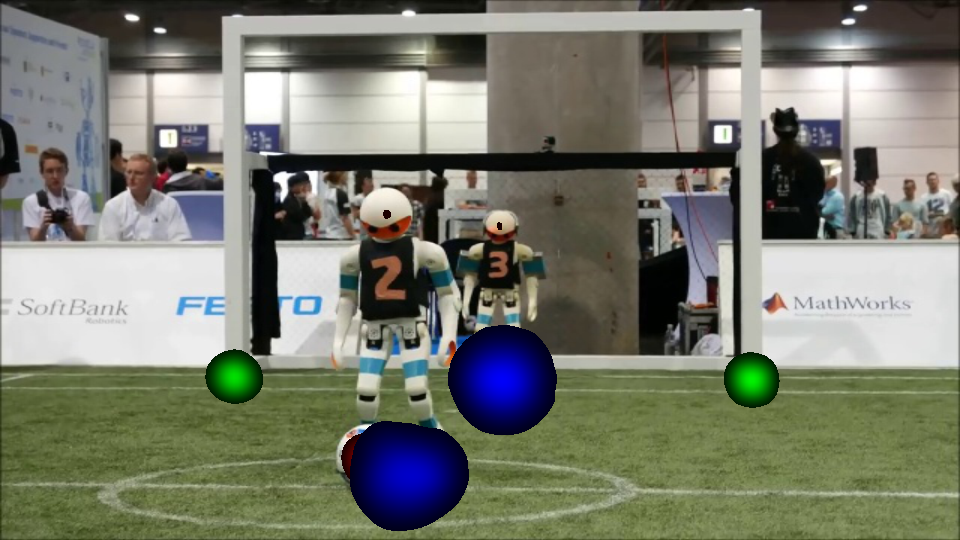}\,%
	\includegraphics[width=.245\textwidth]{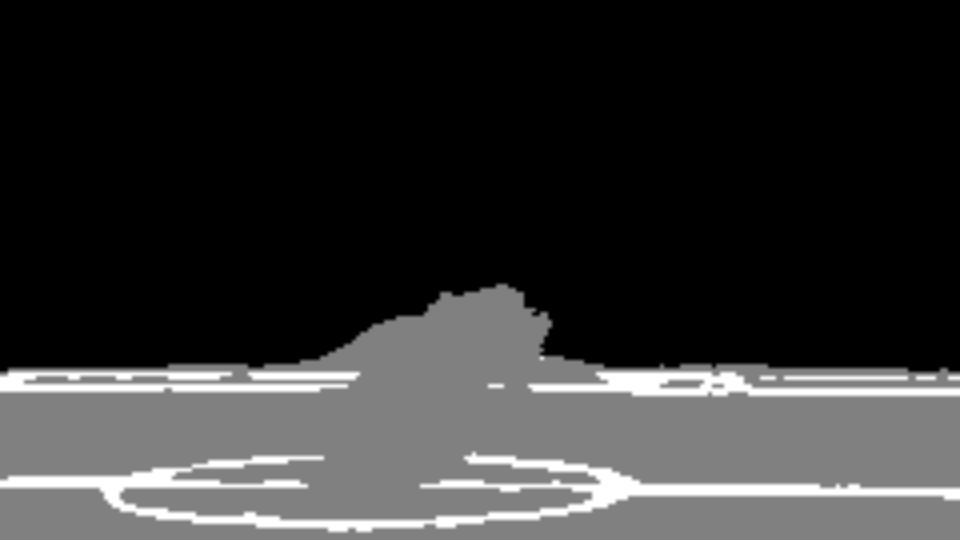}\,%
	\includegraphics[width=.245\textwidth]{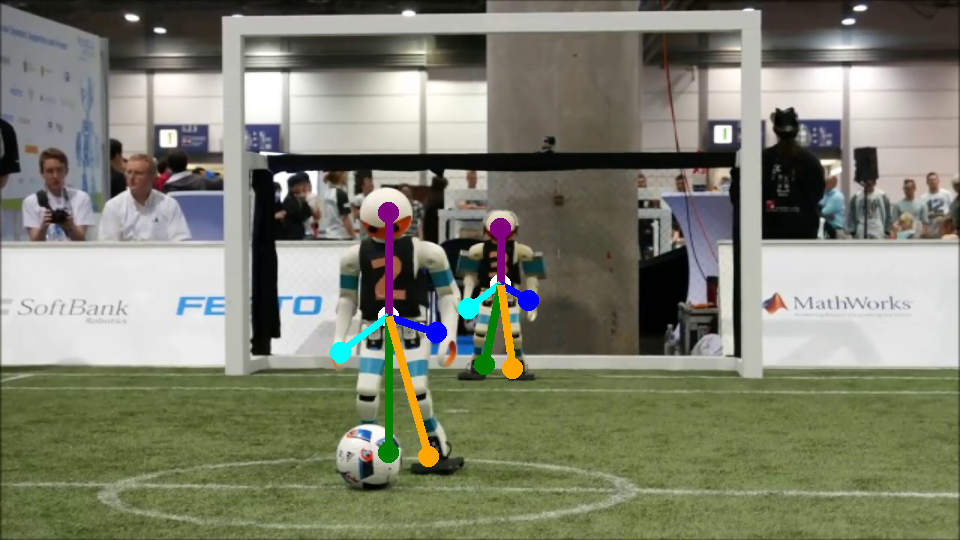}\vspace*{.1ex}\\
	\includegraphics[width=.245\textwidth]{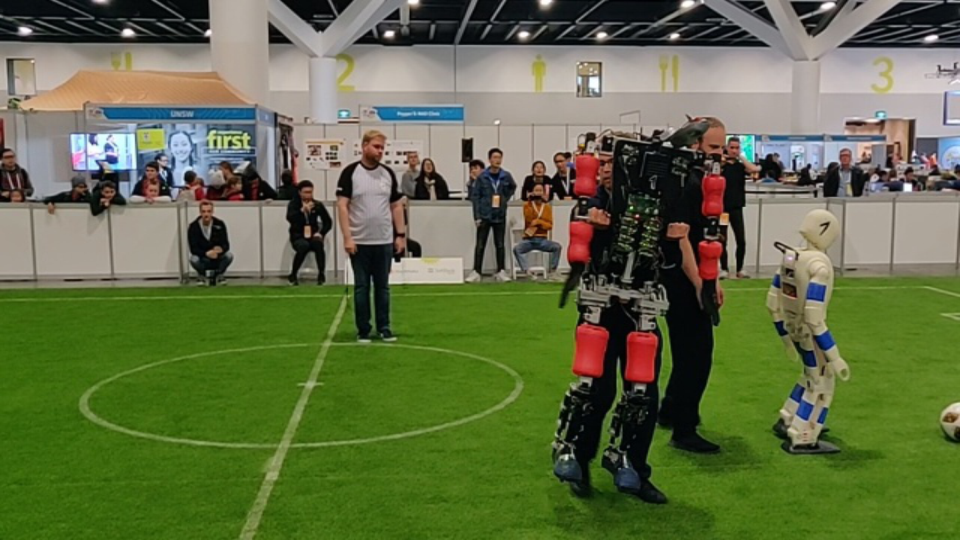}\,%
	\includegraphics[width=.245\textwidth]{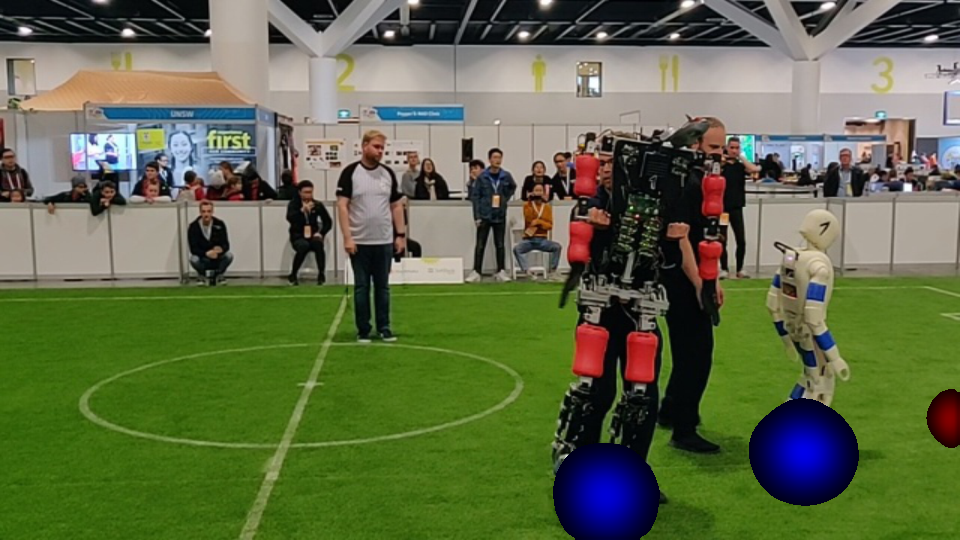}\,%
	\includegraphics[width=.245\textwidth]{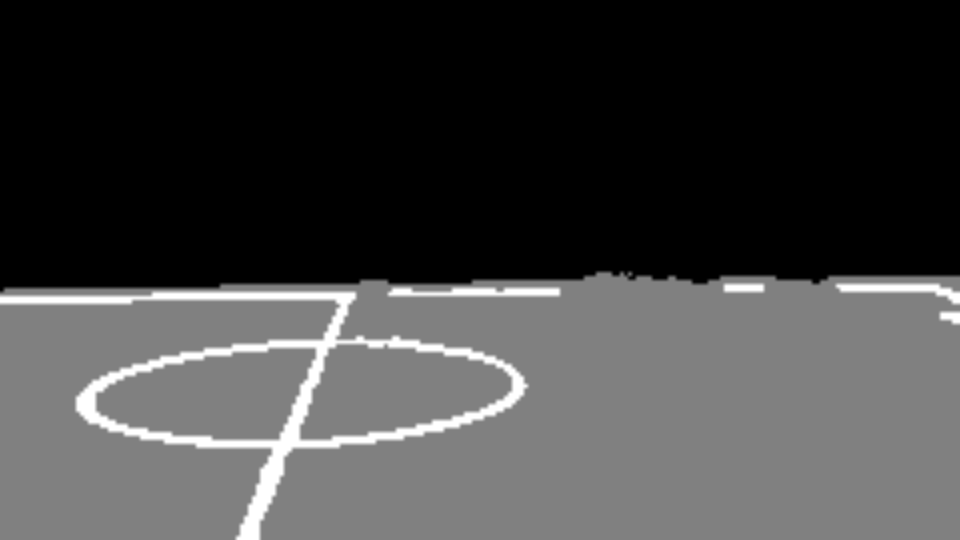}\,%
	\includegraphics[width=.245\textwidth]{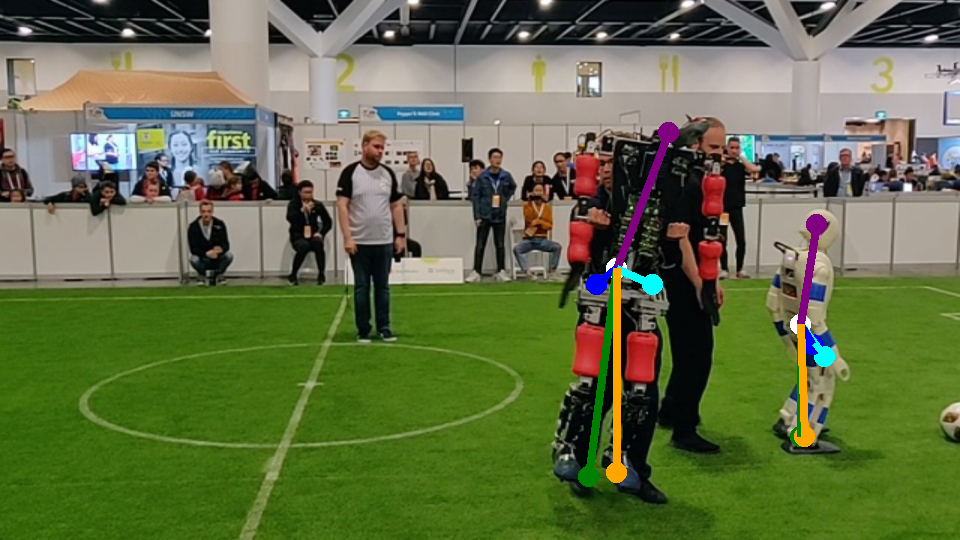}\\\vspace*{.2ex}
  ~~a) Images~~~~~~~~~~~~~~b) Detections~~~~~~~~~~c) Segmentation~~~~~~~~d) Robot Poses\vspace{-1ex}
	\caption{Qualitative results for NimbRoNet3. a) Captured images. b) Predicted object heatmaps for soccer balls (red), robots (blue) and goalposts (green). c) Semantic segmentation of the field with lines (white), field (gray) and background (black). d) Predicted robot poses. We depict the predicted joint locations with a circle and connect all joints corresponding to a single robot.}
	\label{img:preds}
	\vspace*{-4ex}
\end{figure}

Our model is trained to minimize the combined loss function:
\begin{align}
	\Loss & = \lambda_1 \Loss_{Det} + \lambda_2 \Loss_{Seg} + \lambda_3 \Loss_{Pose} + \lambda_4 \Loss_{TV},
\end{align}
where $\Loss_{Det}$ is the mean-squared error between predicted and target object heatmaps, $\Loss_{Seg}$ is the cross entropy loss between predicted and ground-truth segmentation maps, $\Loss_{Pose}$ is the mean-squared error between predicted and target joint keypoints and limb heatmaps, and $\Loss_{TV}$ is a total variation regularization loss that enforces the model to predict smooth heatmaps. The hyper-parameters $\lambda_{1, \ldots,4}$ weight the loss terms.

We train and evaluate our models with a dataset consisting of around 8000 images with annotations for object detection, 1100 samples with ground truth segmentation masks, and 1150 images with annotated robot poses; which are divided $80/20$ for training and testing.
Our data annotation process follows a semi-automatic approach. First, a small set of images was manually annotated using existing annotation tools\footnote{\url{https://supervise.ly
}, \url{https://github.com/bit-bots/imagetagger}}. Then, a trained NimbRoNet3 model is used to generate pseudo-labels for the remaining images, and the user simply has to manually correct those labels that contain errors.
This semi-automatic annotation process allowed us to quickly gather and annotate data of unseen robots and game balls during the RoboCup 2023 competition.

Qualitative results obtained by NimbRoNet3 on test images are shown in Figure~\ref{img:preds}.
Our model simultaneously detects the soccer-related objects, accurately segments the soccer field, and estimates the pose of diverse robots. Additionally, we see how our model generalizes to different viewpoints, lighting conditions, ball designs, and robot appearances.

Our improved perception pipeline allows our humanoid robots to accurately perceive the game environment in real time, being able to detect relevant objects across the soccer field, segment the field lines and boundaries, and to estimate the pose of the opponent robots.
In the future, we plan to employ the estimated robot poses to develop advanced soccer behaviors, such as recognizing and anticipating the actions from our opponents~\cite{ActionRecognition}.

\section{Core Motion Components for Walking and Kicking}

The soccer skills presented by our robots are all combined into a framework,
which extends our previous approaches~\cite{nimbro_winners_2019,nimbro_winners_2022}.
The following subsections summarize all of the necessary features for robust and flexible dynamic walking and in-walk kicking that 
contributed to the dynamic soccer play demonstrated in the RoboCup 2023 Humanoid League AdultSize soccer tournament.

\vspace*{-1ex}
\subsection{Gait Motion Generation}
\vspace*{-0.5ex}
The core of our walking scheme remained unchanged to the previous years~\cite{nimbro_winners_2019}~\cite{nimbro_winners_2022} and revolves around a Central Pattern
Generator~(CPG). Joint trajectories are computed with inverse kinematics from 
end-effector positions, obtained by waveforms designed in the \textit{abstract space} 
and progressing with the \textit{gait phase}.

The \textit{abstract space} is a convenient partitioning of humanoid movement into 
meaningful components e.g. limb swinging, shifting and extension as 
presented in~\cite{Behnke2006} and further extended in~\cite{Allgeuer2016a}. By manually tuning such easily quantifiable 
variables, we obtain a \textit{self-stable} and omnidirectional gait, to which
feedforward and feedback mechanisms are added for rejecting disturbances.

As the approach is model-less and the quantities of the abstract space are bounded, 
we do not need to artificially keep a certain height of the hips or Center of Mass~(CoM).
In fact, we allow our NimbRo-OP2(X) robots to walk with nearly fully extended legs, 
which increases their stride length and simultaneously reduces the necessary knee torque.

\vspace*{-1ex}
\subsection{State Estimation}
\vspace*{-.5ex}
Despite limited sensing capabilities of the NimbRo-OP2(X) robots, we are still 
able of sufficiently reconstructing the robot's state to close the control loop.
By using the kinematic model, absolute joint encoder readings, and torso orientation, 
we obtain the robot's current spatial configuration. 

The torso orientation
is estimated with a non-linear passive complementary filter and represented
in the form of \textit{Fused Angles}~\cite{allgeuer2015fused}. Using \textit{Fused Angles},
we can split the tilt of the robot into uniform sagittal and lateral components,
and apply PID-like feedback on tilt deviations. 

Furthermore, by knowing that the 
soccer field is flat, we assume the lower leg to be the supporting one. To prevent rapid 
support exchanges and assess how much support is provided by each of the legs, we 
apply a height hysteresis. Finally, we track the whole state of a pseudo CoM 
$\mathbf{c} = \begin{bmatrix}c &\dot{c} &\ddot{c}\end{bmatrix}$
in both sagittal and lateral planes. The pseudo CoM position is calculated as a fixed
offset from the torso frame, when the robot is in the nominal gait pose. Due to the
symmetry of the limb movement, this is a sufficiently accurate approximation
of where the actual CoM is. This is then fused in a Kalman filter~\cite{ficht2023direct}
with $\ddot{c}$---the unrotated trunk acceleration measured with the IMU, with gravity $g$ removed---
and used to provide a pseudo Zero Moment Point~(ZMP) $\widehat{p}_z$: \vspace*{-1ex}
\begin{equation} \eqnlabel{pseudozmp}
\widehat{p}_z = \widehat{c} - \frac{\widehat{c}_z}{g}~\widehat{\ddot{c}}.
\end{equation}

\vspace*{-3ex}
\subsection{Feedforward and Feedback Control}
\vspace*{-.5ex}

Before RoboCup 2023, our feedback scheme
was based on the Capture Step Framework~\cite{Missura:CaptureSteps}, which
fitted a Linear Inverted Pendulum Model~(LIPM) to the observed robot movement 
using the CPG-based gait. Using the observed state of the fitted LIPM, 
step-timing and step-size feedback based on a predicted state would then be 
determined and applied to the robot, thus closing the loop. 

Despite its impressive push rejection capabilities, it has a caveat in the form
of decreased responsiveness to commands. As the approach welcomes compliant 
low-gain control to smoothly dissipate a certain level of disturbances, it requires 
overly exaggerated commands to achieve desired effects. This essentially equates to
embedding non-descriptive feedforward compensation for both the lower-level joint 
trajectories and higher-level gait commands. In consequence, the actuators
are always actively tracking a moving target, which results in increased power 
consumption and generated heat.

A unified feedforward controller on the joint level could potentially alleviate 
this problem, but it introduces non-linear behavior which decreases predictability 
by disrupting the initial assumption of the robot behaving like an LIPM.

For the 2023 competition, we extended our Fused Angle Feedback Mechanisms 
gait, previously used in the 2019 competition~\cite{nimbro_winners_2019} to
achieve a similar level of balancing capabilities without compromising 
responsiveness. A feedforward compensation scheme is included, integrating 
RBDL-based Inverse Dynamics~\cite{Felis2016} and a servo-model, which compensates 
for joint torque, friction and supply voltage~\cite{Schwarz2013a}. Due to the
combination of multiple actuators, and external gearing, the model coefficients
had to be manually adjusted to improve joint position tracking.
The gait was also extended with two feedback mechanisms:\vspace*{0.5ex}\\
$\circ$\,\textit{Swing-leg feedback:} Tilt-dependent offsets are applied to the 
abstract swing-leg swing angle (e.g. tilting the robot forward induces
a forward swing of the non-supporting leg, to ensure that the foot lands 
approximately at the desired location with respect to the CoM). The offsets 
are faded out linearly just after the leg begins to transition into the 
supporting leg. As such, the offset is applied only during the respective 
expected swing phase. The scheme is repeated if the tilt progresses into the 
next step.\vspace*{0.5ex}\\
$\circ$\,\textit{Pseudo-CoM-ZMP feedback:} Despite the lack of a model controlling the 
movement of the robot, it still physically behaves like an inverted pendulum 
during its gait cycle and pivots its CoM around the Center of Pressure~(CoP). 
As the open-loop trajectories were designed to be self-stable, we experimentally
observed that the pseudo CoM and pseudo ZMP do in fact approximate the LIPM.
On this basis, we employ a CoM-ZMP controller~\cite{choi2007posture} acting on 
the pseudo states, which realizes input-to-state stability by adjusting $\dot{c}$
to steer $c$ to the reference value. \vspace*{0.5ex}

To generate the reference pseudo CoM and pseudo ZMP values, a second CPG generates 
purely reference motions and assumes nominal CPG-execution and upright torso orientation. 
We match the reference pseudo ZMP location to the observed one on the robot
by adjusting an offset from the center of the supporting foot. 

We adopted the Direct Centroidal Control~(DCC) concept~\cite{ficht2023direct} 
to incorporate leaky integration and applied the integrated pseudo CoM-offset 
to shift the hips of the robot, replacing the fused angle hip-shift feedback. 
Additionally, a high-pass rate-limiting scheme~\cite{englsberger2018torque} 
was used to provide smooth velocity changes, necessary for the inverse dynamics 
calculation in the feedforward compensation.

\vspace*{-1ex}
\subsection{Flexible Waveform-based Kicking}
\vspace*{-.5ex}

\begin{figure}[!b]
	\centering
	\includegraphics[height=38mm]{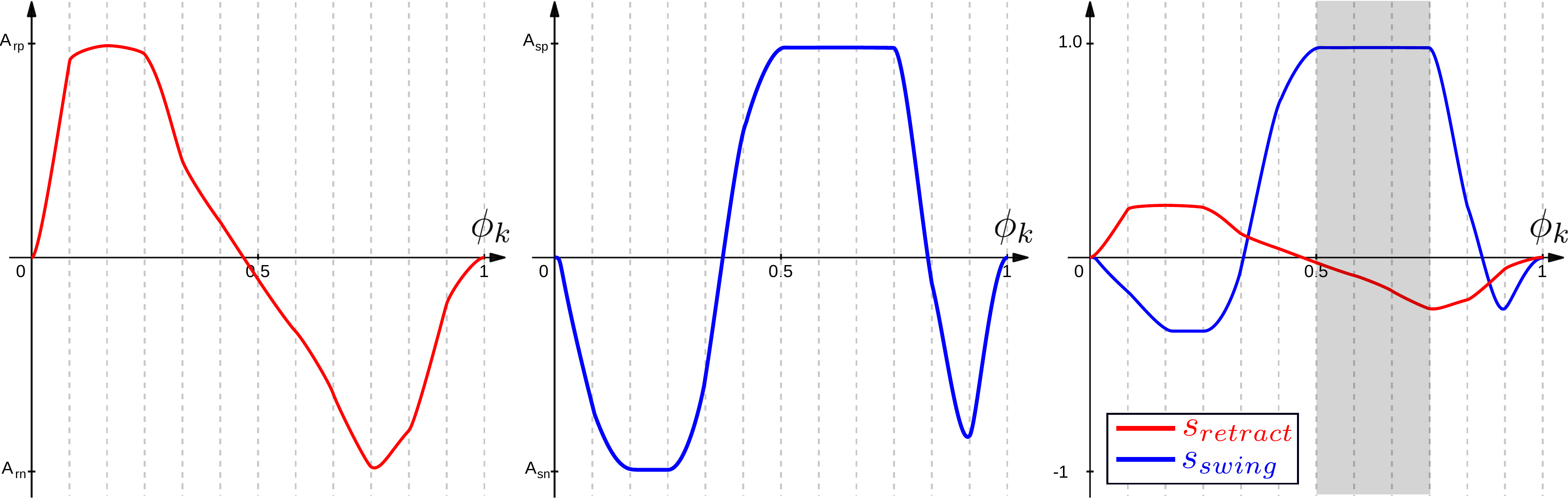}
	\vspace*{-1ex}
	\caption{In-walk kick waveforms. 
	Left: Isolated design of the retraction waveform. Middle: Isolated design of the swing waveform. 
	Right: Absolute applied values with scaling to the abstract gait pose.
	$\phi_k$ is the kick phase (like $x_K$ in Fig. 6 of \cite{nimbro_winners_2019}). 
	$s_{swing}$ and $s_{retract}$ are the experimentally achieved swing and retract waveforms, the gray area
	depicts the time-window where momentum is the highest, e.g. leg is being fully extended and swung forward.}
	\label{fig:kick-wave}
	\vspace{-4ex}
\end{figure}

For RoboCup 2019 in Sydney, we introduced the in-walk kicking technique~\cite{nimbro_winners_2019} integrated
seamlessly into the gait. This was a subtle yet effective change that greatly increased our performance and 
made the game pace more dynamic. In-walk kicking was extended in 2022, where the kicks became more robust to the
alignment of the robot and allowed for directional kicking~\cite{nimbro_winners_2022}. 

However, the distance achieved with each kick would vary largely, barely reaching \SI{4}{m}.
Sometimes four to five kicks would be necessary to score a goal, which gave opponents more opportunities to take ball possession. Having stronger kicks would not only reduce the number of necessary contacts and make goal attempts quicker, but also help clear the ball from the goal in the attempts of opponents. 

To achieve stronger kicks, we implemented a waveform designer to modify the kick parameters on-line.
Two waveforms designed in the abstract space responsible for leg retraction and the swing angle 
are superimposed with the CPG gait trajectories. The waveform editor allows for modifying 12 keypoint values
scaled by the positive and negative maximum amplitudes of the allowed retraction and angle.
This allows for a resolution of approx. \SI{30}{ms} with our current step duration.
The keypoints are linearly interpolated and low-pass filtered for smooth transitions.
With the kicking waveforms shown in~\figref{kick-wave}, we aimed to maximize the momentum transmitted 
to the ball during contact, which happens when both the thigh and shank reach their maximum velocity.

Designing the waveforms on-line allowed for quick iteration of the kicks directly on the robot. The operator can then
quickly iterate and receive feedback on the attempt, to the point where a single 1-hour manual tuning session 
was enough to exceed the previous approach. Tuning directly on the robot ensured that the kicks would be feasible
on the physical system by striking a balance between kick strength and stability, omitting simulation efforts entirely. 
By parameterizing the positive and 
negative amplitudes of the waveforms, we can vary the kick strength depending on the game situation.
The presented approach effectively doubled our kicking distance from an average of \SI{2.5}{m} to \SI{5}{m}, and 
quite often surpassing \SI{7}{m} allowing us to even score goals from our own half.

\section{Higher-level Game Control and Soccer Behaviors}
\seclabel{behaviors}

The decision process for playing soccer is referred to as soccer behaviors. In our system, this functionality is realized by two finite state machines (FSM). 

The Game FSM is responsible for high-level game-related behaviors, such as going to a start position when the game is about to begin, awaiting the opponent to perform a throw-in, etc. The Game FSM is mostly affected by the game state provided by the game controller. 

The Behavior FSM defines soccer skills, such as looking for the ball, approaching the ball, kicking the ball, avoiding obstacles, communicating with teammates, and deciding who will handle the ball depending on the perceived game situation. Please refer to~\cite{ficht2018Grown} for a more detailed description of our soccer behaviors module.

\section{Technical Challenges}
\seclabel{technical_challenges}

Technical Challenges are a separate competition in the RoboCup AdultSize League where the robots compete in performing four predefined tasks in a 25 minute timeslot. The limited time available forces the teams to design robust and reliable solutions. In this section, we present our approaches to the three challenges: Push Recovery, High Kick, and Goal Kick from Moving Ball. 

\subsection{Push Recovery}
\seclabel{push_recovery}

\begin{figure}[t!]
	\centering
 	\includegraphics[width=0.245\linewidth]{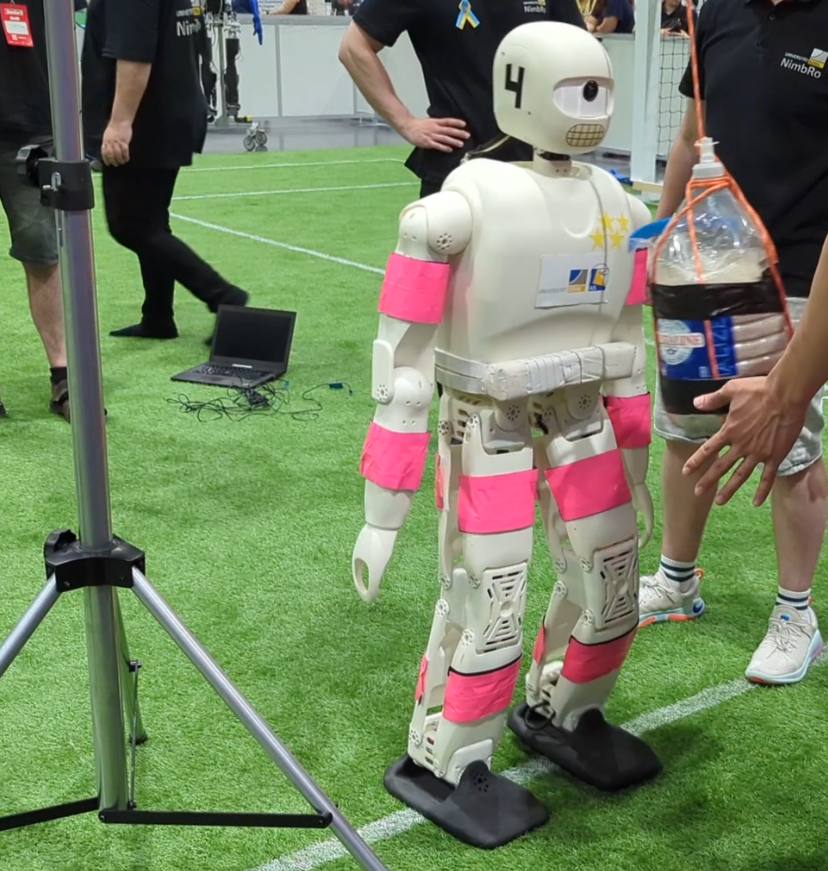}\,%
 	\includegraphics[width=0.245\linewidth]{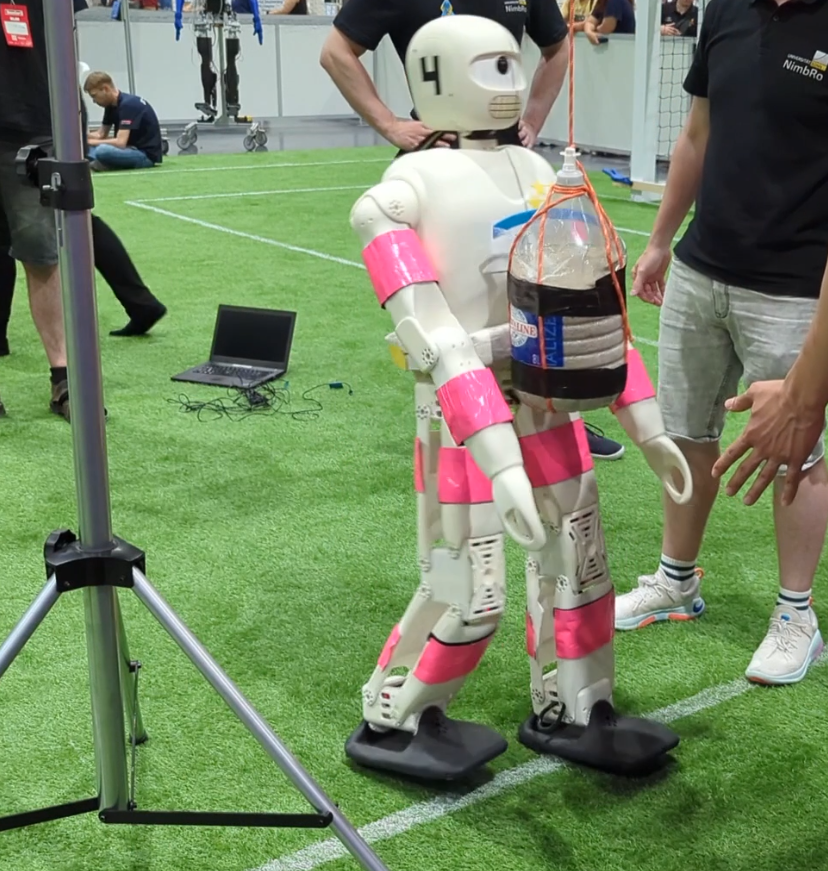}\,%
 	\includegraphics[width=0.245\linewidth]{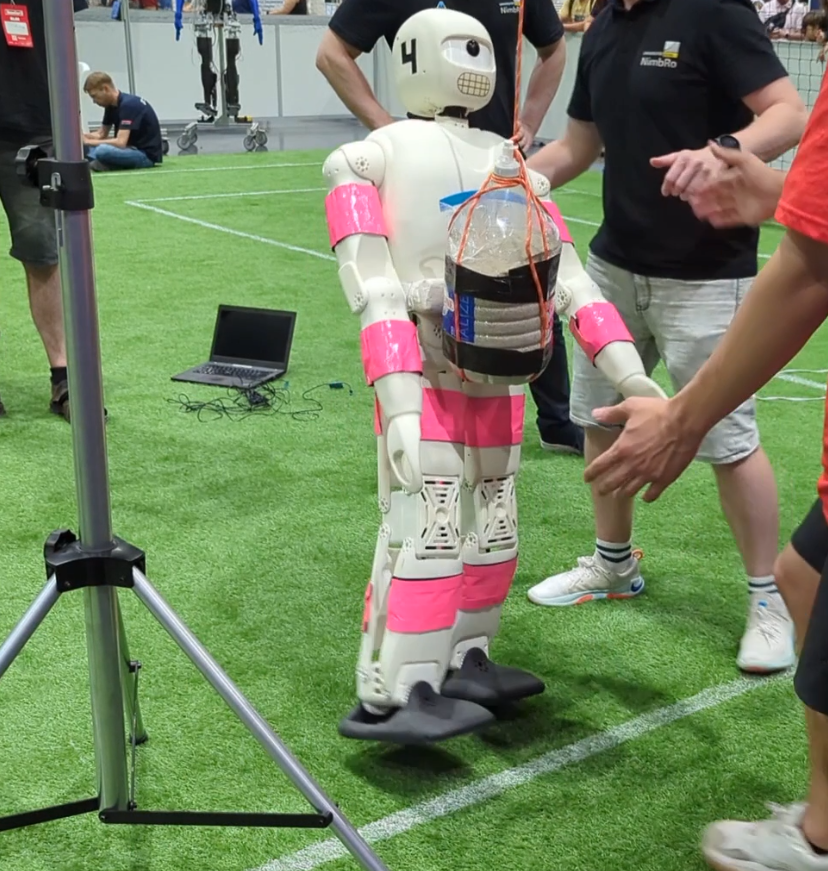}\,%
 	\includegraphics[width=0.245\linewidth]{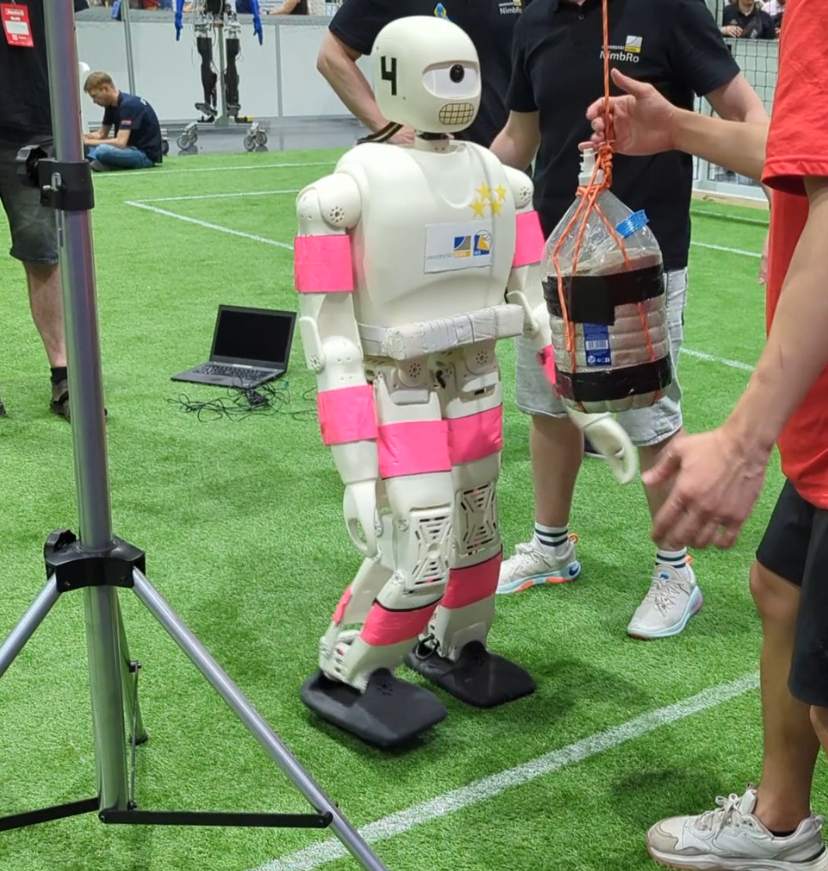}
	\caption{Technical Challenge: Push Recovery. The robot successfully recovers from a frontal push of \SI{10}{kg} pendulum, which is more than half of the robot's weight.}
	\label{fig:pushrecovery}
	\vspace*{-4ex}
\end{figure}

The Push Recovery challenge is the most relevant for Humanoid Soccer, as it tests the robot's 
ability to withstand a set of three pushes while walking on the spot: from the front, from the back, and any of the two. Each push is applied by a free-falling pendulum which impacts the robot at its CoM. The final score achieved depends on how far the pendulum was retracted and the ratio of its mass to that of the robot. For this challenge we used a novel gait, which unifies centroidal state estimation, feedforward and feedback control techniques through a five-mass representation of a humanoid robot~\cite{ficht2020fast}. 

For the NimbRo-OP2X it was particularly important to compensate for the limitations arising from the 4-bar parallel linkage mechanism in the legs, and generate tilt-based step-feedback~\cite{ficht2023centroidal}. Using this approach, the robot completed several trials with a \SI{3}{kg} and \SI{5}{kg} pendulum. Our robot won the challenge by sustaining pushes from a \SI{5}{kg} pendulum, let go at a horizontal distance of \SI{90}{cm}, more than doubling the result of the second-best team. This also superseded our results obtained previously with the Capture Steps Framework. NimbRo-OP2X was also able to withstand two consecutive pushes from a \SI{10}{kg} pendulum~(at~\SI{40}{cm}), which is more than half of its own weight~(see \figref{pushrecovery}).

\begin{figure}[b!]
	\centering
	\includegraphics[width=0.245\linewidth]{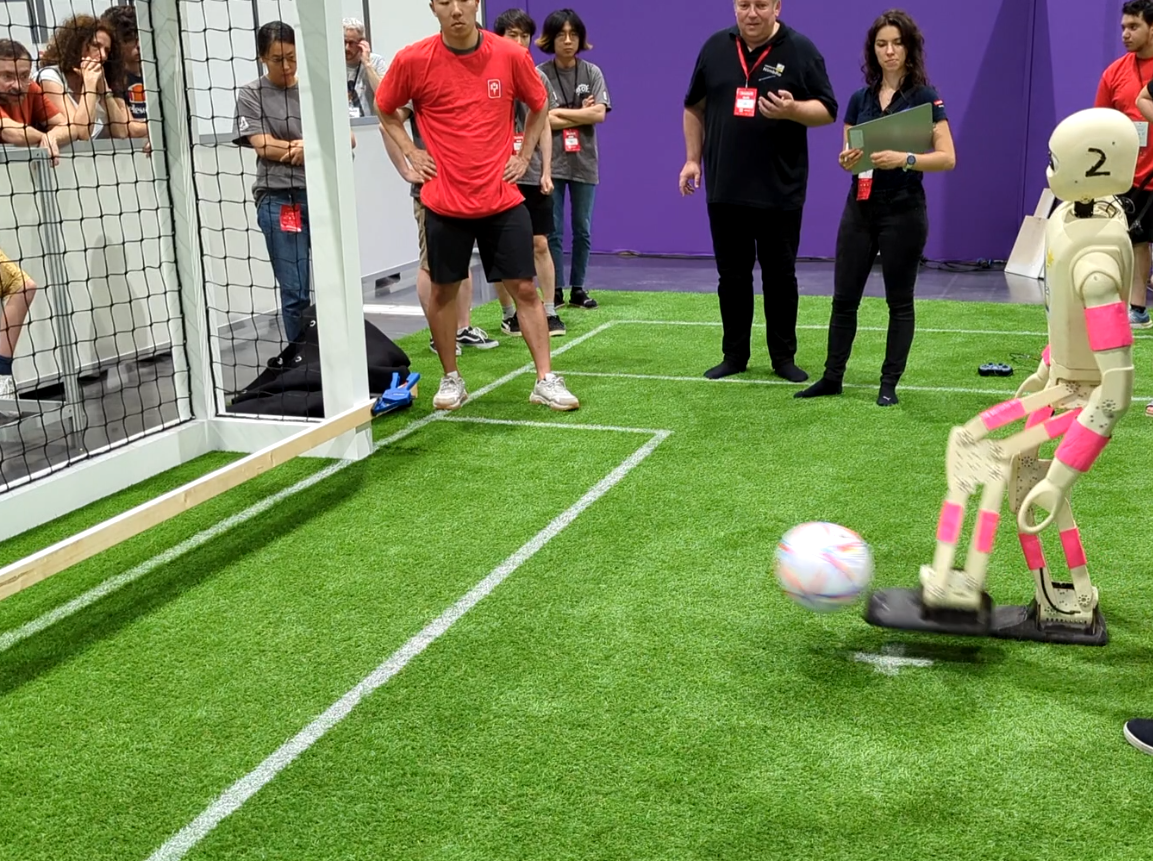}\,%
	\includegraphics[width=0.245\linewidth]{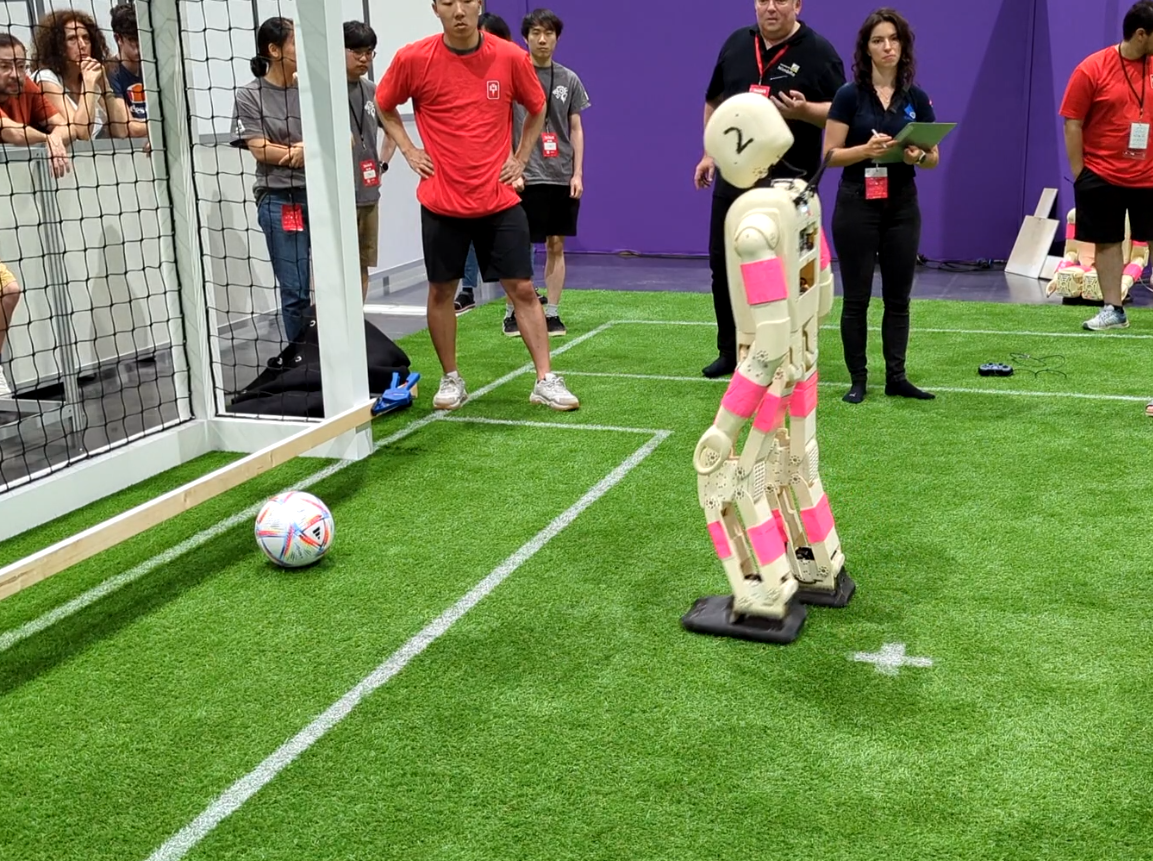}\,%
	\includegraphics[width=0.245\linewidth]{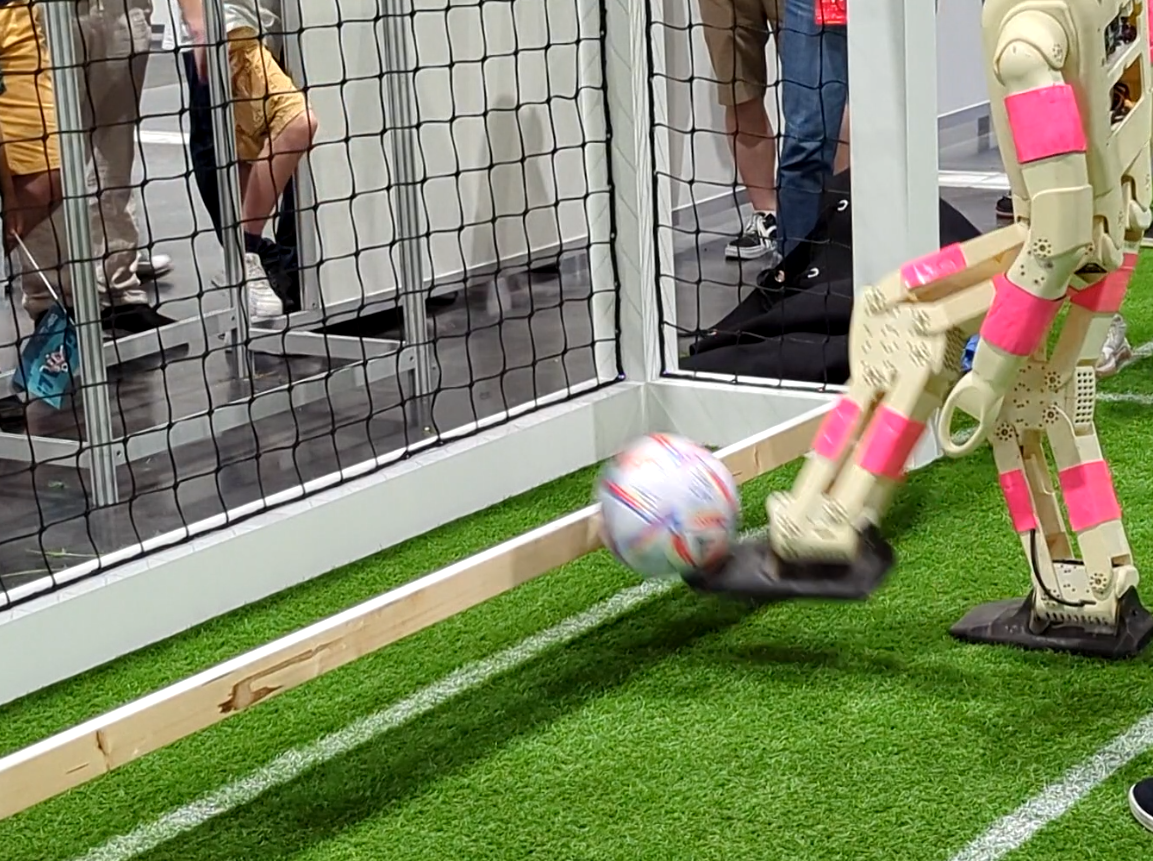}\,%
	\includegraphics[width=0.245\linewidth]{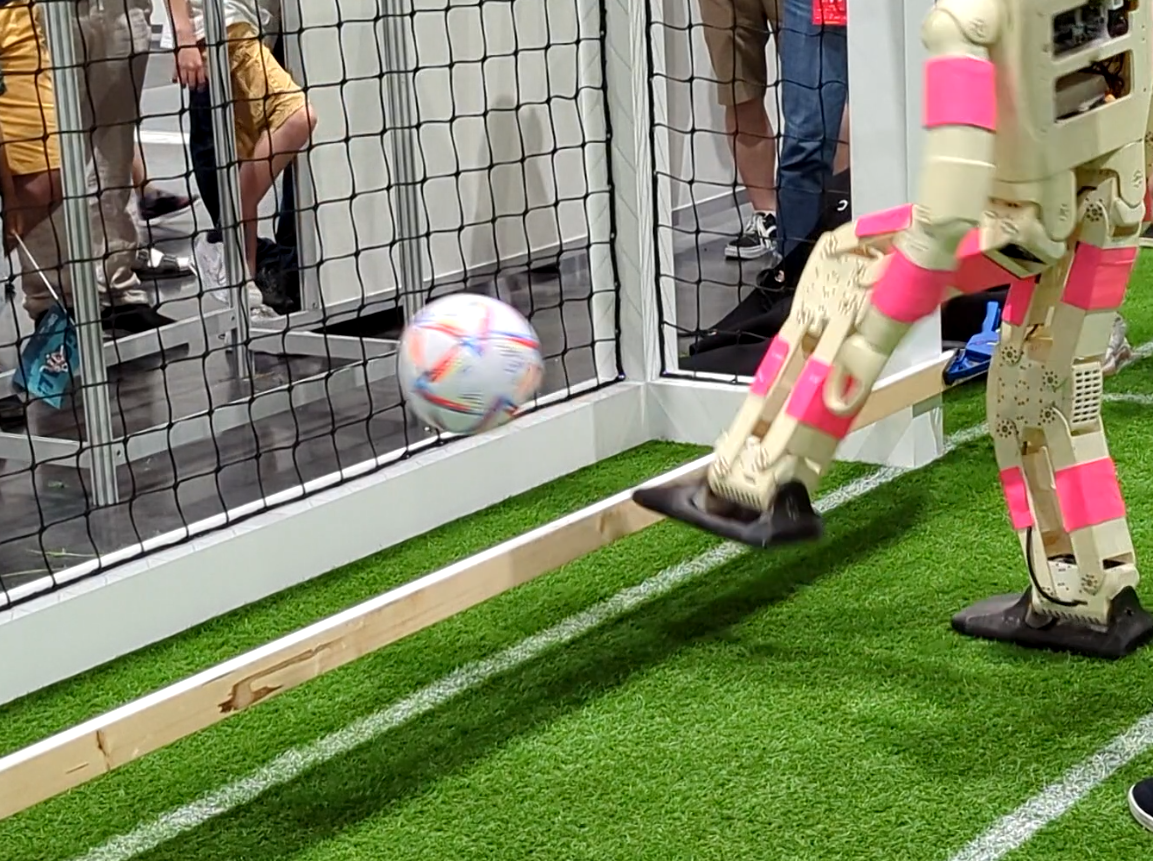}
	\caption{Technical Challenge: High Kick. First, the robot moves the ball closer to the obstacle by making a weak kick. Second, the robot approaches the re-positioned ball and kicks it over the obstacle.}
	\label{fig:high_kick}
	\vspace*{-4ex}
\end{figure}

\vspace*{-1ex}
\subsection{High Kick}
\seclabel{high_kick}
\vspace*{-.5ex}

The objective of this challenge is to score a goal such that the ball flies over an obstacle placed at the goal line. The challenge starts with the ball on the penalty mark. To maximize the height which we can kick over, we first move the ball closer to the obstacle by performing a weak kick, defined as a predesigned motion. After this, we approach the ball and perform a predesigned high kick motion, which aims at kicking the ball at a closest-to-the-ground point, thus directing the ball upward. Our robot successfully scored a goal over a 20\,cm obstacle (\figref{high_kick}), coming in second in this challenge. Team HERoEHS won this challenge, kicking over a 21\,cm obstacle.

\begin{figure}[t!]
	\centering
	\includegraphics[width=0.245\linewidth]{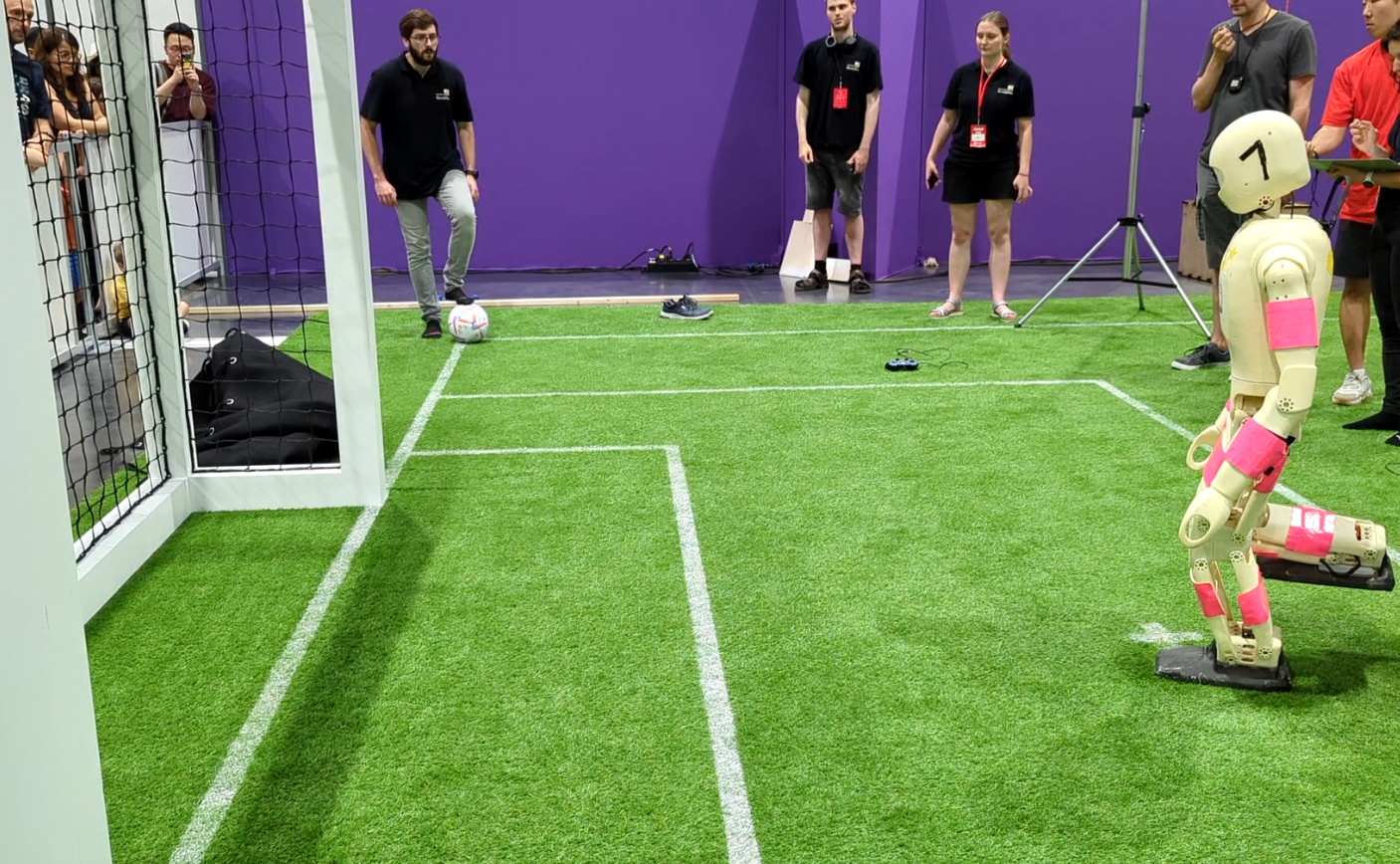}\,%
	\includegraphics[width=0.245\linewidth]{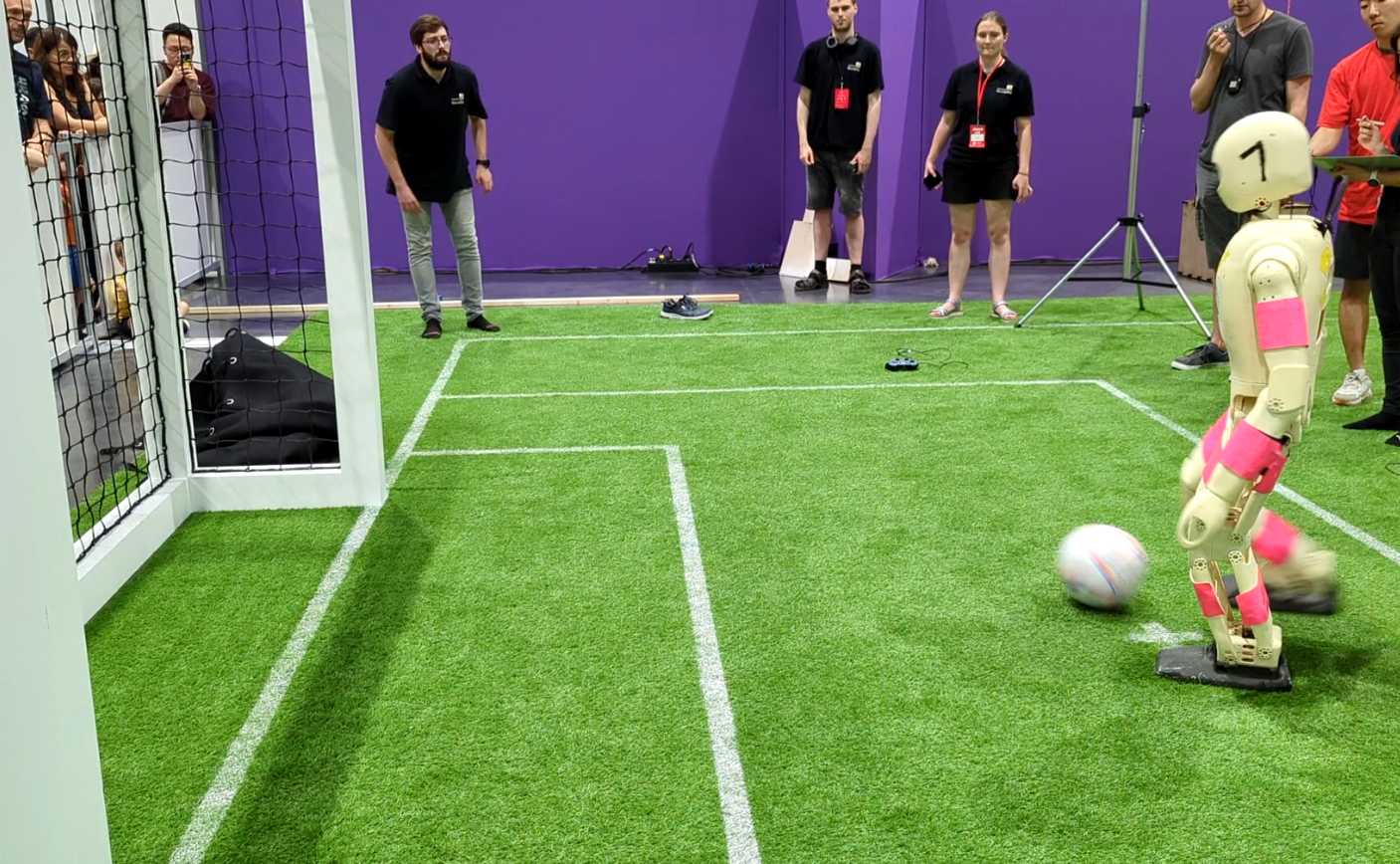}\,%
	\includegraphics[width=0.245\linewidth]{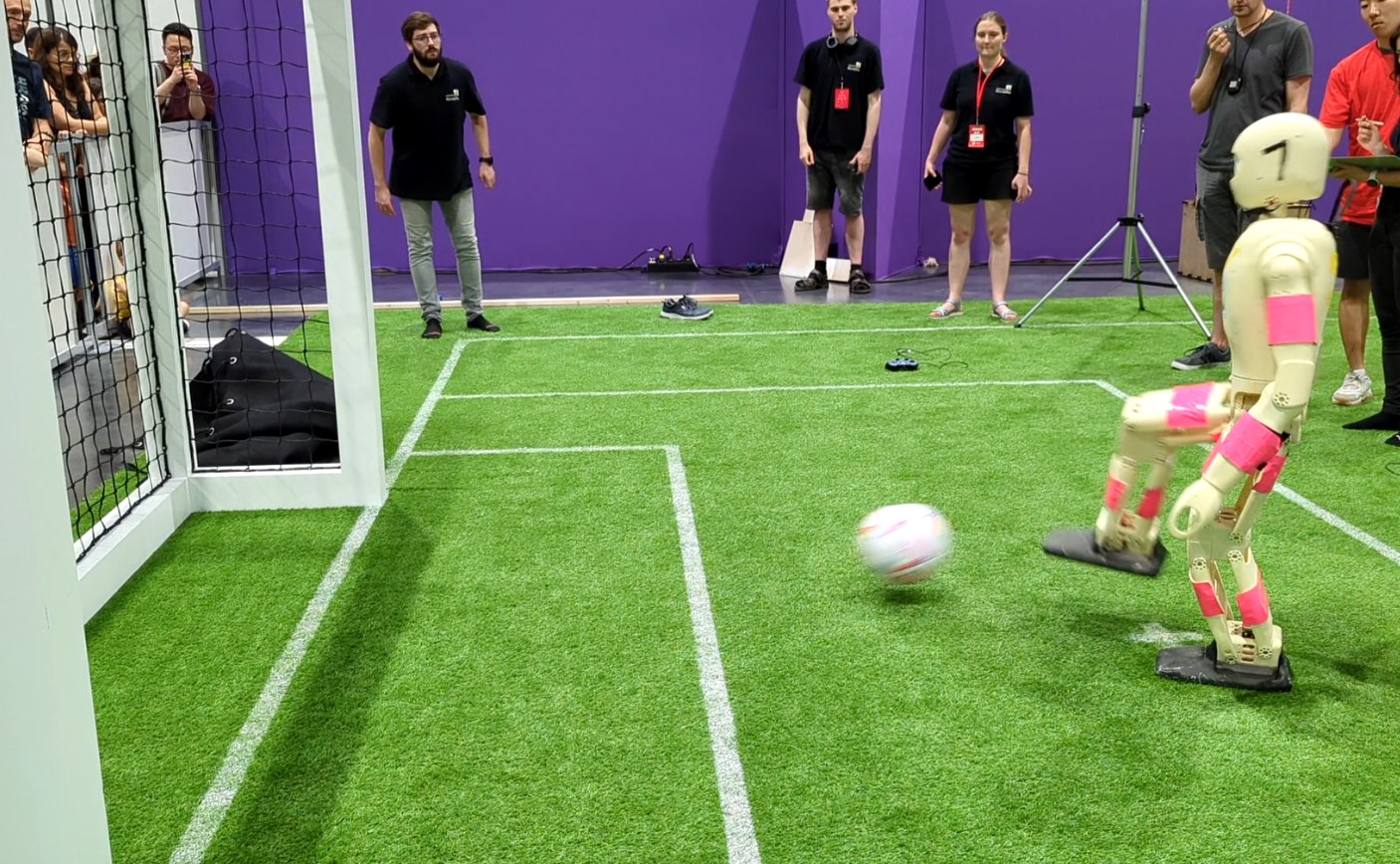}\,%
	\includegraphics[width=0.245\linewidth]{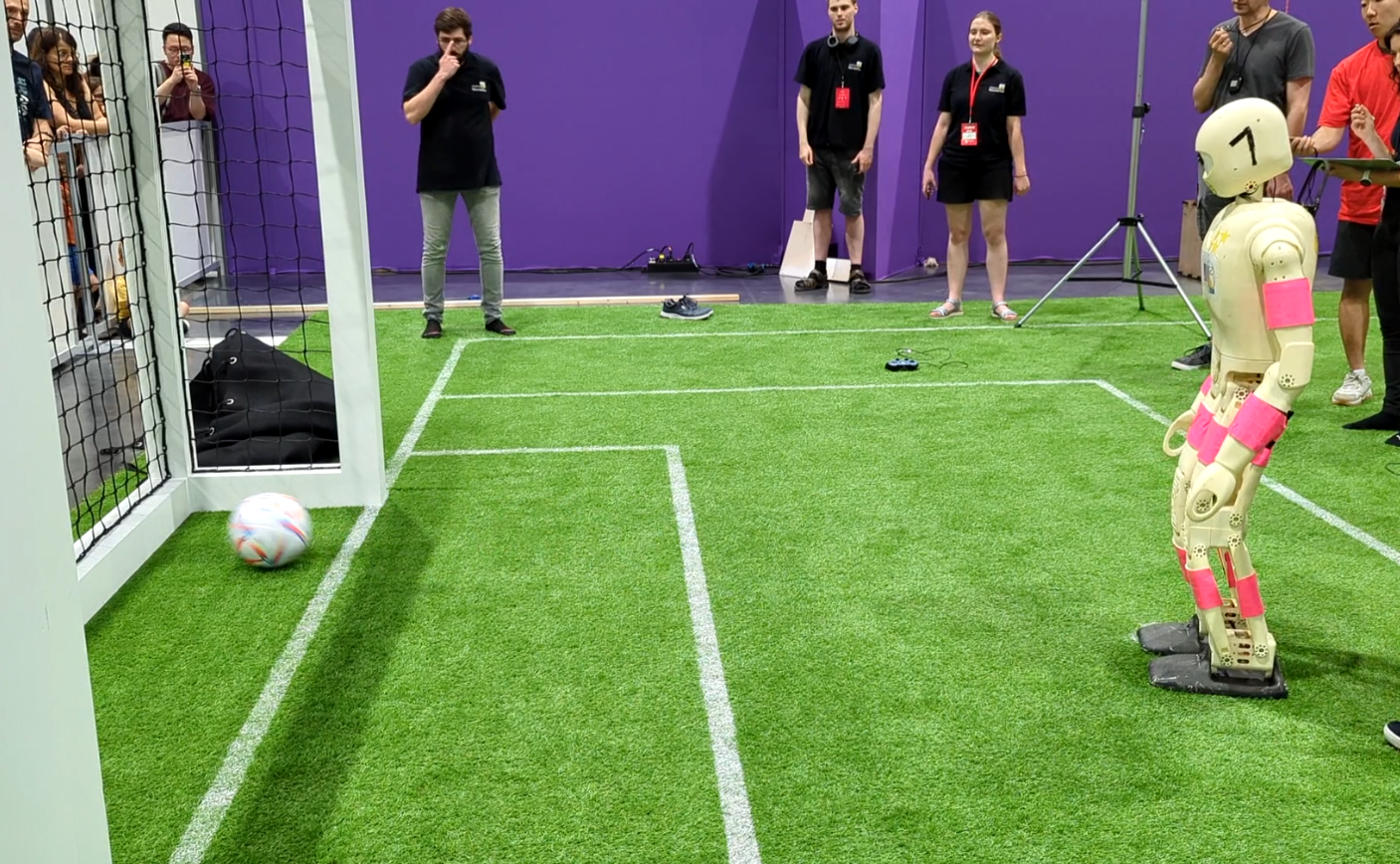}
	\caption{Technical Challenge: Goal Kick from Moving Ball.}
	\label{fig:moving_ball}
	\vspace*{-4ex}
\end{figure}

\vspace*{-1ex}
\subsection{Goal Kick from Moving Ball}
\seclabel{moving_ball}
\vspace*{-.5ex}

The objective of this challenge is to score a goal by kicking a ball which is moving towards the robot. The robot starts at the penalty mark and the pass is performed by a human. To reliably kick at the right moment, we use ball detections from our visual perception module to estimate its velocity and acceleration, which then enables us to estimate the time of arrival to the region in front of the foot. Finally, knowing the time needed for a kick motion to be executed, we identify the moment when the kick has to be triggered. Our robot managed to score two goals out of three passes (\figref{moving_ball}), coming in second in this challenge. Team HERoEHS won this challenge, scoring three goals in a row.

\section{Soccer Game Performance}

At the RoboCup 2023 AdultSize soccer competition, teams consisting of two autonomous humanoid robots played matches against each other. This year, four teams participated in the AdultSize League. 

Prior to the main soccer tournament, a Drop-In competition, where robots from different teams had to collaborate during 3 vs. 3 matches, took place. In this competition, our robots played four games with a cumulative score of 26:0. All these goals were scored by our robots. 

During the round-robin of the main soccer tournament, our robots won all their three matches with a total score of 22:0. In the semifinals, our team played against team RoMeLa from UCLA (USA), winning the match with a score of 10:0. In the final, we met team HERoEHS from Hanyang University (Korea). Our robots played well and won the final game with the score of 8:0, becoming the champions in the AdultSize League for the fifth year in a row. Overall, our robots demonstrated a solid performance throughout all the matches they played without losing a single goal, achieving a total score of 66:0. 

\section{Conclusions}

In this paper, we presented the improvements to our previous-year RoboCup-winning humanoid soccer system. 
The improvements enabled us to win again the main soccer tournament as well as Drop-In competition in the Humanoid AdultSize League at RoboCup 2023 in Bordeaux. 

In particular, we introduced an updated visual perception model NimbRoNet3 capable of estimating poses of the opponent robots. Further, we proposed an extended fused angles feedback mechanism for footstep adjustment and an additional COM-ZMP controller. These updates improved walking stability which contributed to our robots staying in the game without falling for extended periods of time, including episodes of intense whole-body contacts with much heavier robots. The introduced waveform-based formulation for parametric in-walk kicks enabled our robots to reliably score goals from greater distances, including goals from our own half of the field. Finally, a newly introduced gait helped us to win the Push Recovery technical challenge, outperforming our results from previous years.

% Acknowledgements
\subsection*{Acknowledgments}
\footnotesize
This work was partially funded by H2020 project EUROBENCH, GA 779963.

% Bibliography
\bibliographystyle{splncs03}
\bibliography{winners_2023}

\end{document}